\documentclass{article}

\usepackage[final]{nips_2016}
\usepackage[utf8]{inputenc}
\usepackage[T1]{fontenc}
\usepackage{url}
\usepackage{nicefrac}
\usepackage{microtype}
\usepackage[export]{adjustbox}
\usepackage{mwhmacros}
\usepackage{xcolor,colortbl}
\usepackage{amsmath}
\usepackage{bm}
\usepackage{wrapfig}
\usepackage{titlesec}

\title{Learning to learn by gradient descent\\\hspace{7.75em}by gradient descent}

\author{
Marcin Andrychowicz$^1$,
Misha Denil$^1$,
Sergio G\'omez Colmenarejo$^1$,
Matthew W. Hoffman$^1$,\\
\textbf{David Pfau$^1$,
Tom Schaul$^1$,
Brendan Shillingford$^{1,2}$,
Nando de Freitas$^{1,2,3}$}\\[0.5em]
$^1$Google DeepMind \quad $^2$University of Oxford \quad $^3$Canadian Institute for Advanced Research \\[0.5em]
\texttt{marcin.andrychowicz@gmail.com}\\
\texttt{\{mdenil,sergomez,mwhoffman,pfau,schaul\}@google.com}\\
\texttt{brendan.shillingford@cs.ox.ac.uk},
\texttt{nandodefreitas@google.com}
}

\begin{document}
\maketitle


\begin{abstract}
  The move from hand-designed features to learned features in machine learning
  has been wildly successful.  In spite of this, optimization algorithms are
  still designed by hand.  In this paper we show how the design of an
  optimization algorithm can be cast as a learning problem, allowing the
  algorithm to learn to exploit structure in the problems of interest in an
  automatic way.  Our learned algorithms, implemented by LSTMs, outperform
  generic, hand-designed competitors on the tasks for which they are trained,
  and also generalize well to new tasks with similar structure.  We demonstrate
  this on a number of tasks, including simple convex problems, training neural
  networks, and styling images with neural art.
\end{abstract}

\section{Introduction}

Frequently, tasks in machine learning can be expressed as the problem of
optimizing an objective function $f(\theta)$ defined over some domain
$\theta\in\Theta$. The goal in this case is to find the minimizer
$\theta^* =
\argmin_{\theta\in\Theta}f(\theta)$. While any method capable of minimizing
this objective function can be applied, the standard approach for
differentiable functions is some form of gradient descent, resulting in a
sequence of updates
\begin{align*}
  \theta_{t+1} = \theta_t - \alpha_t \nabla f(\theta_t)
  \,.
\end{align*}
The performance of vanilla gradient descent, however, is hampered by the fact
that it \emph{only} makes use of gradients and ignores second-order
information. Classical optimization techniques correct this behavior by
rescaling the gradient step using curvature information, typically via the
Hessian matrix of
second-order partial derivatives---although other choices such as the
generalized Gauss-Newton matrix or Fisher information matrix are possible.

Much of the modern work in optimization is based around designing update rules
tailored to specific classes of problems, with the types of problems of
interest differing between different research communities.  For example, in the
deep learning community we have seen a proliferation of optimization methods
specialized for high-dimensional, non-convex optimization problems. These
include momentum \citep{nesterov:1983,tseng:1998}, Rprop \citep{riedmiller:1992},
Adagrad \citep{duchi:2011}, RMSprop \citep{tieleman:2012}, and ADAM \citep{kingma:2015}. More
focused methods can also be applied when more structure of the optimization
problem is known
\citep{martens:2015}. In contrast, communities who focus on sparsity tend to
favor very different approaches~\citep{donoho:2006,bach:2012}. This is even
more the case for combinatorial optimization for which relaxations are often
the norm \citep{nemhauser:1988}.

\begin{wrapfigure}[18]{r}{0.5\textwidth}
\centering
\includegraphics[width=\linewidth]{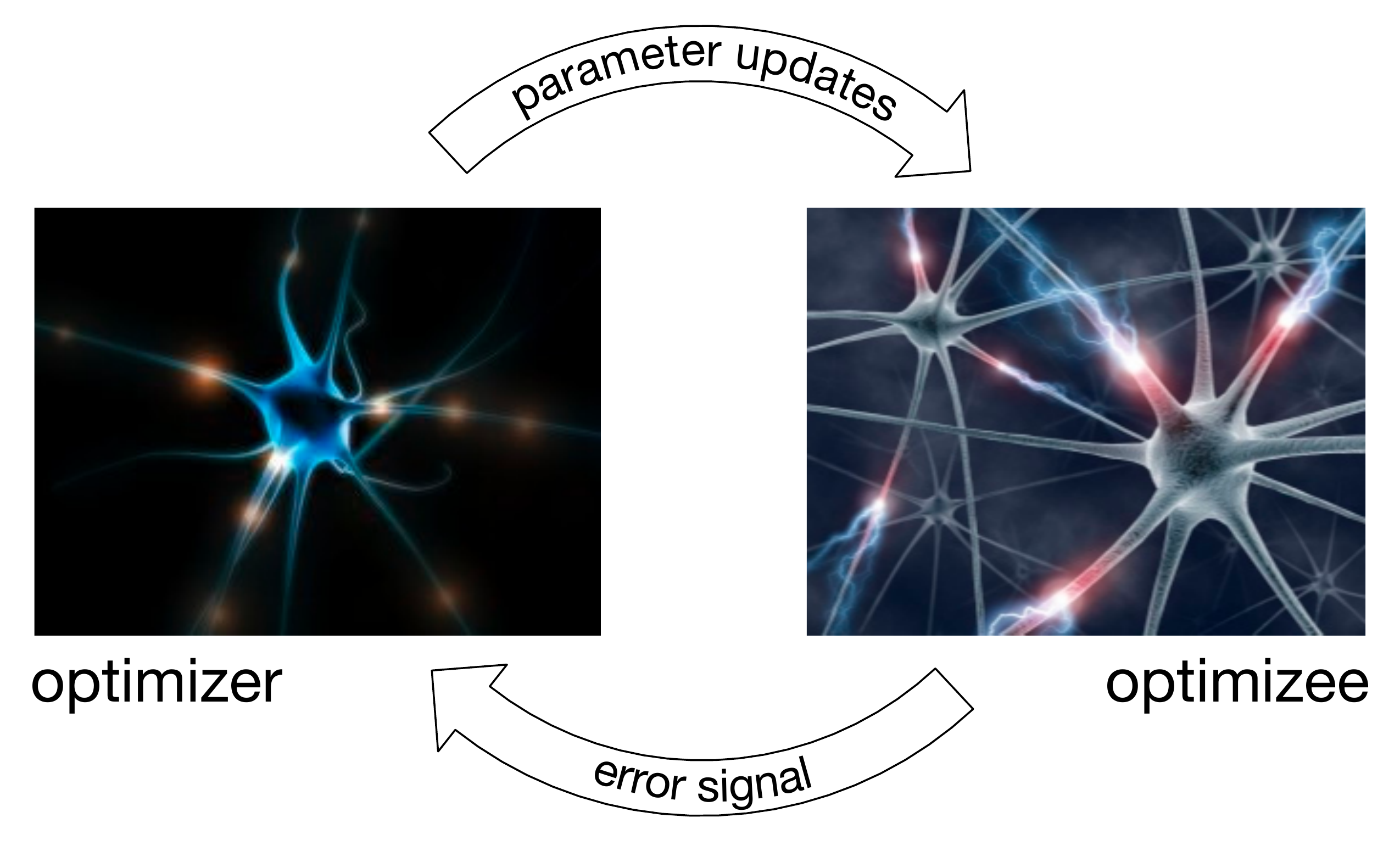}
\caption{The optimizer (left) is provided with performance of the optimizee
(right) and proposes updates to increase the optimizee's performance.
\citep[photos:][]{neuron1,neuron2}}
\label{fig:optimizee}
\end{wrapfigure}

This industry of optimizer design allows different communities to create
optimization methods which exploit structure in their problems of interest at
the expense of potentially poor performance on problems outside of that scope.
Moreover the \emph{No Free Lunch Theorems for Optimization}
\citep{wolpert:1997} show that in the setting of combinatorial optimization, no
algorithm is able to do better than a random strategy in expectation.  This
suggests that specialization to a subclass of problems is in fact the
\emph{only} way that improved performance can be achieved in general.


In this work we take a different tack and instead propose to replace hand-designed update rules with a learned update rule, which we call the optimizer
$g$, specified by its own set of parameters $\phi$. This results in updates to
the optimizee $f$ of the form
\begin{equation}
  \theta_{t+1} = \theta_t + g_t(\nabla f(\theta_t), \phi)
  \,.
  \label{eq:meta-update}
\end{equation}
A high level view of this process is shown in Figure~\ref{fig:optimizee}.  In
what follows we will explicitly model the update rule $g$ using a recurrent
neural network (RNN) which maintains its own state and hence dynamically
updates as a function of its iterates.



\subsection{Transfer learning and generalization}

The goal of this work is to develop a procedure for constructing a learning
algorithm which performs well on a particular class of optimization problems.
Casting algorithm design as a learning problem allows us to specify the class
of problems we are interested in through example problem instances.  This is in
contrast to the ordinary approach of characterizing properties of interesting
problems analytically and using these analytical insights to design learning
algorithms by hand.

It is informative to consider the meaning of \emph{generalization} in this
framework.  In ordinary statistical learning we have a particular function of
interest, whose behavior is constrained through a data set of example function
evaluations.  In choosing a model we specify a set of inductive biases about
how we think the function of interest should behave at points we have not
observed, and generalization corresponds to the capacity to make predictions
about the behavior of the target function at novel points.
In our setting the examples are themselves \emph{problem instances}, which
means generalization corresponds to the ability to transfer knowledge between
different problems.  This reuse of problem structure is commonly known as
\emph{transfer learning}, and is often treated as a subject in its own right.
However, by taking a meta-learning perspective, we can cast the problem of
transfer learning as one of generalization, which is much better studied in the
machine learning community.

One of the great success stories of deep-learning is that we can rely on the
ability of deep networks to generalize to new examples by learning interesting
sub-structures. In this work we aim to leverage this generalization power, but
also to lift it from simple supervised learning to the more general setting of
optimization.

\subsection{A brief history and related work}

The idea of using \emph{learning to learn} or \emph{meta-learning} to acquire
knowledge or inductive biases has a long history \citep{thrun:1998}. More
recently, \cite{lake:2016} have argued forcefully for its importance as a
building block in artificial intelligence. Similarly, \cite{santoro:2016}
frame multi-task learning as generalization, however unlike our approach they
directly train a base learner rather than a training algorithm.  In general
these ideas involve learning which occurs at two different time scales: rapid
learning within tasks and more gradual, \emph{meta} learning across many
different tasks.

Perhaps the most general approach to meta-learning is that of
\cite{schmidhuber:1992, schmidhuber:1993}---building on work from
\citep{schmidhuber:1987}---which considers networks that are able to modify
their own weights. Such a system is differentiable end-to-end, allowing both
the network and the learning algorithm to be trained jointly by gradient
descent with few restrictions.  However this generality comes at the expense of
making the learning rules very difficult to train.  Alternatively, the work of
\cite{schmidhuber:1997} uses the Success Story Algorithm to modify its search
strategy rather than gradient descent; a similar approach has been recently
taken in \cite{daniel:2016} which uses reinforcement learning to train a
controller for selecting step-sizes.

\cite{bengio:1990, bengio:1995} propose to learn updates which avoid
back-propagation by using simple parametric rules. In relation to the focus of
this paper the work of \citeauthor{bengio:1990} could be characterized as
\emph{learning to learn \textbf{without} gradient descent by gradient descent}.
The work of \cite{runarsson:2000} builds upon this work by replacing the simple
rule with a neural network.

\cite{cotter:1990}, and later \cite{younger:1999}, also show fixed-weight
recurrent neural networks can exhibit dynamic behavior without need to modify
their network weights. Similarly this has been shown in a filtering context
\citep[e.g.][]{feldkamp:1998}, which is directly related to simple
multi-timescale optimizers \citep{sutton:1992, schraudolph:1999}.

Finally, the work of \cite{younger:2001} and \cite{hochreiter:2001} connects
these different threads of research by allowing for the output of
backpropagation from one network to feed into an additional \emph{learning}
network, with both networks trained jointly. Our approach to meta-learning
builds on this work by modifying the network architecture of the optimizer in
order to scale this approach to larger neural-network optimization problems.

\section{Learning to learn with recurrent neural networks}
\label{sec:learning-to-learn}


In this work we consider directly parameterizing the optimizer. As a result, in
a slight abuse of notation we will write the final \emph{optimizee parameters}
$\theta^*(f, \phi)$ as a function of the \emph{optimizer parameters} $\phi$ and
the function in question. We can then ask the question: What does it mean for
an optimizer to be good? Given a distribution of functions $f$ we will write
the expected loss as
\begin{equation}
    \calL(\phi) =
    \mathbb{E}_{f} \Big[ f\big(\theta^*(f, \phi)\big) \Big] \,.
    \label{eq:meta-optimization-final}
\end{equation}
As noted earlier, we will take the update steps $g_t$ to be the output of a
recurrent neural network $m$, parameterized by $\phi$, whose
state we will denote explicitly with $h_t$. Next, while the objective function
in \eqref{eq:meta-optimization-final} depends only on the final parameter
value, for training the optimizer it will be convenient to have an objective
that depends on the entire trajectory of optimization, for some horizon T,
\begin{equation}
    \calL(\phi)
    = \mathbb{E}_{f} \left[\sum_{t=1}^T w_t f(\theta_t)\right]
    \qquad\text{where}\qquad
    \begin{aligned}[t]
        \theta_{t+1} &= \theta_t + g_t \,,
        \\
        \begin{bmatrix}
            g_t \\ h_{t+1}
        \end{bmatrix}
        &= m(\nabla_t, h_t, \phi) \,.
    \end{aligned}
    \label{eq:meta-optimization}
\end{equation}
Here $w_t \in \mathbb{R}_{\ge 0}$ are arbitrary weights associated with each
time-step and we will also use the notation $\nabla_t=\nabla_\theta
f(\theta_t)$. This formulation is equivalent to
\eqref{eq:meta-optimization-final} when $w_t = \boldsymbol{1}[t = T]$,
but later we will describe why using different weights can prove useful.

\begin{figure}
\includegraphics[width=1.0\linewidth]{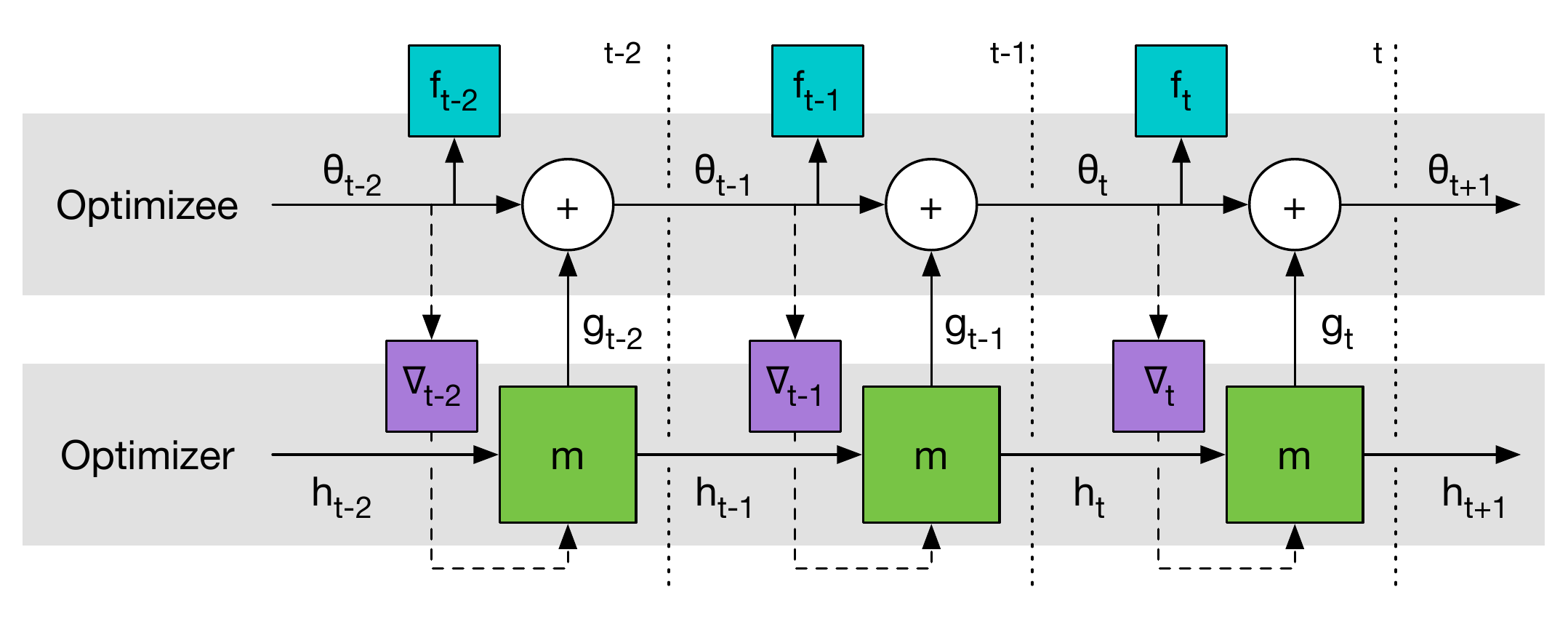}
\caption{Computational graph used for computing the gradient of the optimizer.}
\label{fig:graph}
\end{figure}

We can minimize the value of $\mathcal{L}(\phi)$ using gradient descent on
$\phi$. 
The gradient estimate
$\partial\mathcal{L(\phi)}/\partial\phi$ can be computed by sampling a random
function $f$ and applying backpropagation to the computational graph in
Figure~\ref{fig:graph}.  We allow gradients to flow along the solid edges in
the graph, but gradients along the dashed edges are dropped.  Ignoring
gradients along the dashed edges amounts to making the assumption that the
gradients of the optimizee do not depend on the optimizer parameters, i.e.\
$\partial \nabla_{t}\big/\partial\phi = 0$.
This assumption allows us to avoid computing second
derivatives of $f$.

Examining the objective in \eqref{eq:meta-optimization} we see that
the gradient is non-zero only for terms where $w_t \neq 0$. If we
use $w_t = \boldsymbol{1}[t = T]$ to match the original problem, then gradients
of trajectory prefixes are zero and only the final optimization step
provides information for training the optimizer. This renders
Backpropagation Through Time (BPTT) inefficient.
We solve this problem by relaxing the objective such that $w_t > 0$ at
intermediate points along the trajectory.  This changes the objective function,
but allows us to train the optimizer on partial trajectories.
For simplicity, in all our experiments we use $w_t = 1$ for every $t$.


\subsection{Coordinatewise LSTM optimizer}
\label{sec:coordinatewise}

One challenge in applying RNNs in our setting is that we want to be able to
optimize at least tens of thousands of parameters. Optimizing at this scale with a fully
connected RNN is not feasible as it would require a huge hidden state and an
enormous number of parameters. To avoid this difficulty we will use an
optimizer $m$ which operates coordinatewise on the parameters of the objective
function, similar to other common update rules like RMSprop and ADAM. This
\emph{coordinatewise network architecture} allows us to use a very small network that only looks at a single coordinate to
define the optimizer and share optimizer parameters across different parameters
of the optimizee.

Different behavior on each coordinate is achieved by using separate activations
for each objective function parameter. In addition to allowing us to use a
small network for this optimizer, this setup has the nice effect of making the
optimizer invariant to the order of parameters in the network, since the same
update rule is used independently on each coordinate.

\begin{wrapfigure}{r}{0.5\textwidth}
\centering
\includegraphics[width=\linewidth,clip]{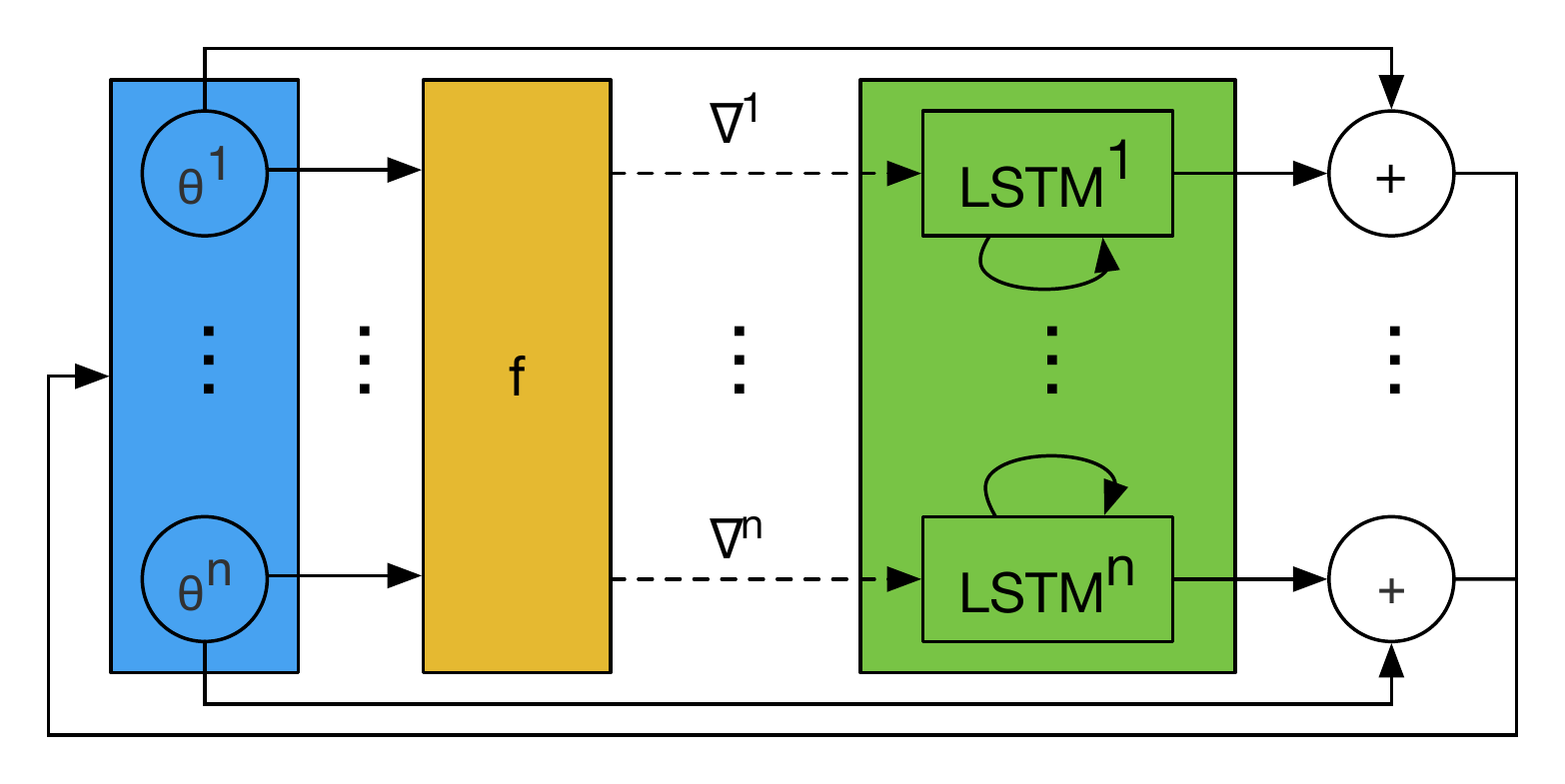}
\caption{One step of an LSTM optimizer. All LSTMs have shared
parameters, but separate hidden states.}
\label{fig:lstm}
\end{wrapfigure}

We implement the update rule for each coordinate using a two-layer Long Short
Term Memory (LSTM) network \citep{hochreiter:1997}, using the now-standard
forget gate architecture. The network takes as input the optimizee gradient for
a single coordinate as well as the previous hidden state and outputs the update
for the corresponding optimizee parameter. We will refer to this architecture,
illustrated in Figure~\ref{fig:lstm}, as an LSTM optimizer.

The use of recurrence allows the LSTM to learn dynamic update rules which
integrate information from the history of gradients, similar to momentum. This
is known to have many desirable properties in convex optimization
\citep[see e.g.][]{nesterov:1983} and in fact many recent learning
procedures---such as ADAM---use momentum in their updates.

\paragraph{Preprocessing and postprocessing}

Optimizer inputs and outputs can have very different magnitudes depending on
the class of function being optimized, but neural networks usually work
robustly only for inputs and outputs which are neither very small nor very
large. In practice rescaling inputs and outputs of an LSTM optimizer using
suitable constants (shared across all timesteps and functions $f$) is
sufficient to avoid this problem. In Appendix~\ref{sec:pre} we propose a
different method of preprocessing inputs to the optimizer inputs which is more
robust and gives slightly better performance.

\section{Experiments}\label{sec:exp}

In all experiments the trained optimizers use two-layer LSTMs with 20 hidden
units in each layer. Each optimizer is trained by minimizing
Equation~\ref{eq:meta-optimization} using truncated BPTT as described in
Section~\ref{sec:learning-to-learn}. The minimization is performed using ADAM
with a learning rate chosen by random search.

We use early stopping when training the optimizer in order to avoid overfitting
the optimizer.  After each epoch (some fixed number of learning steps) we
freeze the optimizer parameters and evaluate its performance.  We pick the best
optimizer (according to the final validation loss) and report its average
performance on a number of freshly sampled test problems.

We compare our trained optimizers with standard optimizers used in Deep
Learning: SGD, RMSprop, ADAM, and Nesterov's accelerated gradient (NAG).
For each of these optimizer and each problem we tuned the learning rate, and
report results with the rate that gives the best final error for each problem.
When an optimizer has more parameters than just a learning rate (e.g. decay
coefficients for ADAM) we use the default values from the \texttt{optim}
package in Torch7.  Initial values of all optimizee parameters were sampled
from an IID Gaussian distribution.

\subsection{Quadratic functions}

In this experiment we consider training an optimizer on a simple class of
synthetic 10-dimensional quadratic functions.  In particular we consider minimizing
functions of the form
\begin{align*}
    f(\theta) = \|W\theta - y\|_2^2
\end{align*}
for different 10x10 matrices $W$ and 10-dimensional vectors $y$ whose elements are drawn
from an IID Gaussian distribution.
Optimizers were trained by optimizing random functions from this family and
tested on newly sampled functions from the same distribution.  Each function was
optimized for 100 steps and the trained optimizers were unrolled for 20 steps.
We have not used any preprocessing, nor postprocessing.

\begin{figure}
\centering
\includegraphics[width=\linewidth]{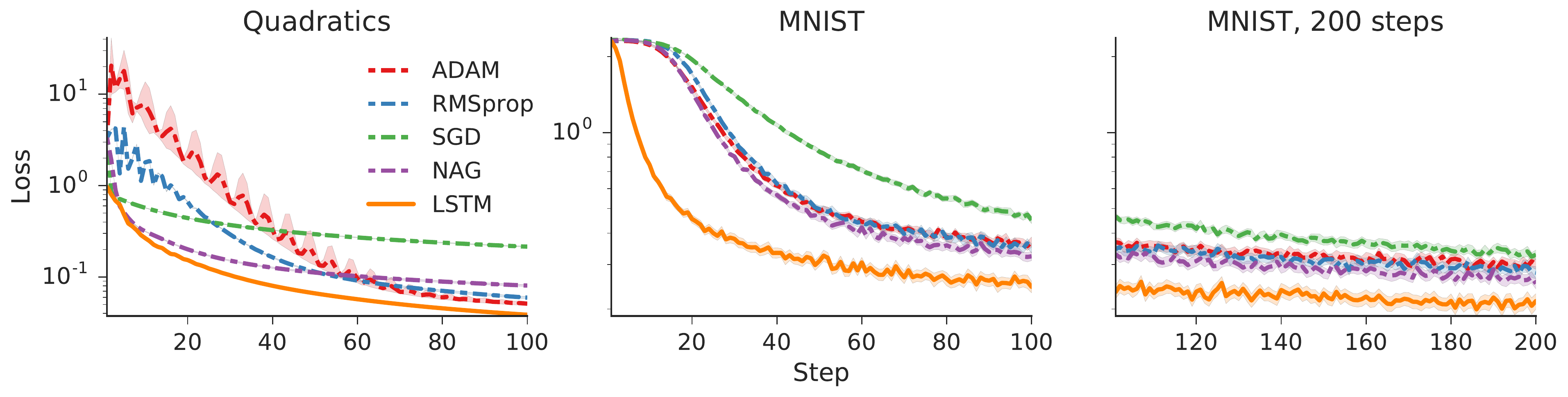}
\caption{Comparisons between learned and hand-crafted optimizers performance.
Learned optimizers are shown with solid lines and hand-crafted optimizers are
shown with dashed lines.  Units for the $y$ axis in the MNIST plots are logits.
\textbf{Left:} Performance of different optimizers on randomly sampled
10-dimensional quadratic functions. \textbf{Center:} the LSTM optimizer
outperforms standard methods training the base network on MNIST.
\textbf{Right:} Learning curves for steps 100-200 by an optimizer trained to
optimize for 100 steps (continuation of center plot).}
\label{fig:mnist-1}
\end{figure}

Learning curves for different optimizers, averaged over many functions, are
shown in the left plot of Figure~\ref{fig:mnist-1}.  Each curve corresponds to
the average performance of one optimization algorithm on many test functions;
the solid curve shows the learned optimizer performance and dashed curves show
the performance of the standard baseline optimizers.  It is clear the learned
optimizers substantially outperform the baselines in this setting.

\begin{figure}
\centering
\includegraphics[width=\linewidth]{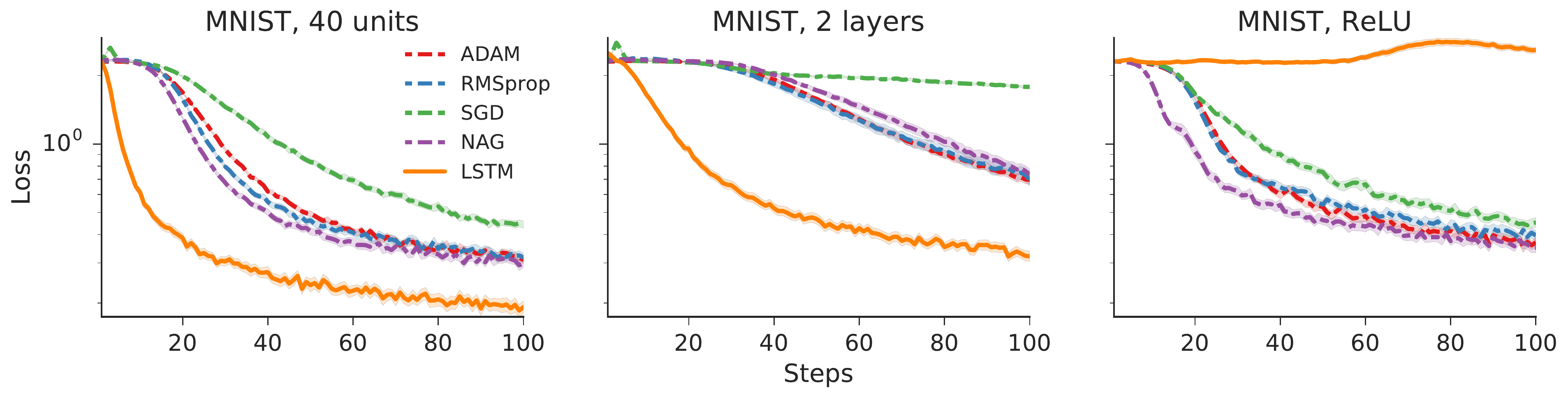}
\caption{Comparisons between learned and hand-crafted optimizers performance.
Units for the $y$ axis are logits. \textbf{Left:} Generalization to the
different number of hidden units (40 instead of 20).  \textbf{Center:}
Generalization to the different number of hidden layers (2 instead of 1).  This
optimization problem is very hard, because the hidden layers are very narrow.
\textbf{Right:} Training curves for an MLP with 20 hidden units using ReLU
activations.  The LSTM optimizer was trained on an MLP with sigmoid
activations.}
\label{fig:mnist-2}
\end{figure}

\begin{figure}
\centering
\includegraphics[width=\linewidth]{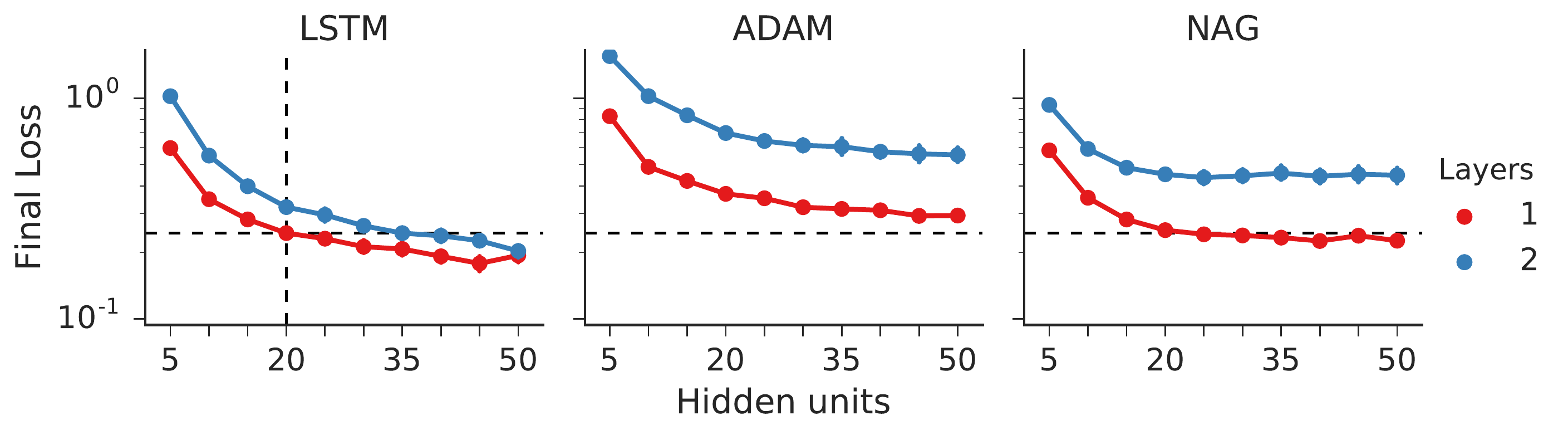}
\caption{Systematic study of final MNIST performance as the optimizee
architecture is varied, using sigmoid non-linearities. The vertical dashed line
in the left-most plot denotes the architecture at which the LSTM is trained and
the horizontal line shows the final performance of the trained optimizer in
this setting.}
\label{fig:mnist-sigmoid}
\end{figure}

\subsection{Training a small neural network on MNIST}

In this experiment we test whether trainable optimizers can learn to optimize a
small neural network on MNIST, and also explore how the trained optimizers
generalize to functions beyond those they were trained on.  To this end, we
train the optimizer to optimize a base network and explore a series of
modifications to the network architecture and training procedure at test time.


In this setting the objective function $f(\theta)$ is the cross entropy of a
small MLP with parameters $\theta$. The values of $f$ as well as the gradients
$\partial f(\theta) / \partial\theta$ are estimated using random minibatches of
128 examples. The base network is an MLP with one hidden layer of 20 units
using a sigmoid activation function. The only source of variability between
different runs is the initial value $\theta_0$ and randomness in minibatch
selection. Each optimization was run for 100 steps and the trained optimizers
were unrolled for 20 steps. We used input preprocessing described in
Appendix~\ref{sec:pre} and rescaled the outputs of the LSTM by the factor
$0.1$.

Learning curves for the base network using different optimizers are displayed
in the center plot of Figure~\ref{fig:mnist-1}. In this experiment NAG, ADAM,
and RMSprop exhibit roughly equivalent performance the LSTM optimizer
outperforms them by a significant margin. The right plot in
Figure~\ref{fig:mnist-1} compares the performance of the LSTM optimizer if it
is allowed to run for 200 steps, despite having been trained to optimize for
100 steps.  In this comparison we re-used the LSTM optimizer from the previous
experiment, and here we see that the LSTM optimizer continues to outperform the
baseline optimizers on this task.

\paragraph{Generalization to different architectures} Figure~\ref{fig:mnist-2}
shows three examples of applying the LSTM optimizer to train networks with
different architectures than the base network on which it was trained.  The
modifications are (from left to right) (1) an MLP with 40 hidden units instead
of 20, (2) a network with two hidden layers instead of one, and (3) a network
using ReLU activations instead of sigmoid.  In the first two cases the LSTM
optimizer generalizes well, and continues to outperform the hand-designed
baselines despite operating outside of its training regime.  However, changing
the activation function to ReLU makes the dynamics of the learning procedure
sufficiently different that the learned optimizer is no longer able to
generalize. Finally, in Figure~\ref{fig:mnist-sigmoid} we show the results of
systematically varying the tested architecture; for the LSTM results we again
used the optimizer trained using 1 layer of 20 units and sigmoid
non-linearities. Note that in this setting where the test-set problems are
similar enough to those in the training set we see even better generalization
than the baseline optimizers.

\subsection{Training a convolutional network on CIFAR-10}

Next we test the performance of the trained neural optimizers on optimizing
classification performance for the CIFAR-10 dataset \citep{krizhevsky:2009}. In
these experiments we used a model with both convolutional and feed-forward
layers. In particular, the model used for these experiments includes three
convolutional layers with max pooling followed by a fully-connected layer with
32 hidden units; all non-linearities were ReLU activations with batch
normalization.

The coordinatewise network decomposition introduced in
Section~\ref{sec:coordinatewise}---and used in the previous
experiment---utilizes a single LSTM architecture with shared weights, but
separate hidden states, for each optimizee parameter. We found that this
decomposition was not sufficient for the model architecture introduced in this
section due to the differences between the fully connected and convolutional
layers. Instead we modify the optimizer by introducing two LSTMs: one proposes
parameter updates for the fully connected layers and the other updates the
convolutional layer parameters. Like the previous LSTM optimizer we still
utilize a coordinatewise decomposition with shared weights and individual
hidden states, however LSTM weights are now shared only between parameters of
the same type (i.e.\ fully-connected vs.\ convolutional).

\begin{figure}
\centering
\includegraphics[width=\linewidth]{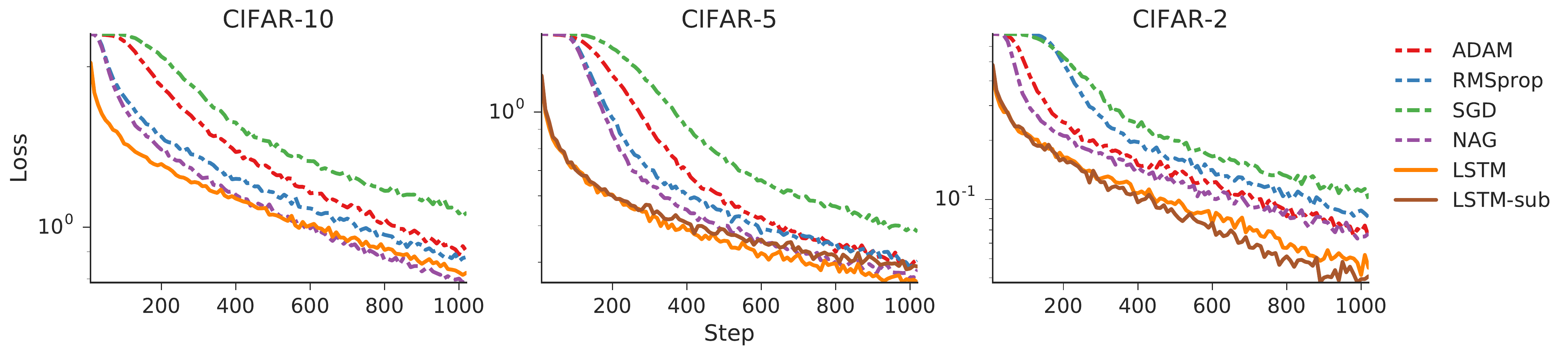}
\caption{Optimization performance on the CIFAR-10 dataset and subsets. Shown on
the left is the LSTM optimizer versus various baselines trained on CIFAR-10 and
tested on a held-out test set. The two plots on the right are the performance
of these optimizers on subsets of the CIFAR labels. The additional optimizer
\emph{LSTM-sub} has been trained only on the heldout labels and is hence
transferring to a completely novel dataset.}
\label{fig:cifar}
\end{figure}

The performance of this trained optimizer compared against the baseline
techniques is shown in Figure~\ref{fig:cifar}. The left-most plot displays the
results of using the optimizer to fit a classifier on a held-out test set. The
additional two plots on the right display the performance of the trained
optimizer on modified datasets which only contain a subset of the labels, i.e.
the CIFAR-2 dataset only contains data corresponding to 2 of the 10 labels.
Additionally we include an optimizer \emph{LSTM-sub} which was only trained on
the held-out labels.

In all these examples we can see that the LSTM optimizer learns much more
quickly than the baseline optimizers, with significant boosts in performance
for the CIFAR-5 and especially CIFAR-2 datsets. We also see that the optimizers
trained only on a disjoint subset of the data is hardly effected by this
difference and transfers well to the additional dataset.

\subsection{Neural Art}

\begin{figure}
\centering
\includegraphics[width=0.85\linewidth]{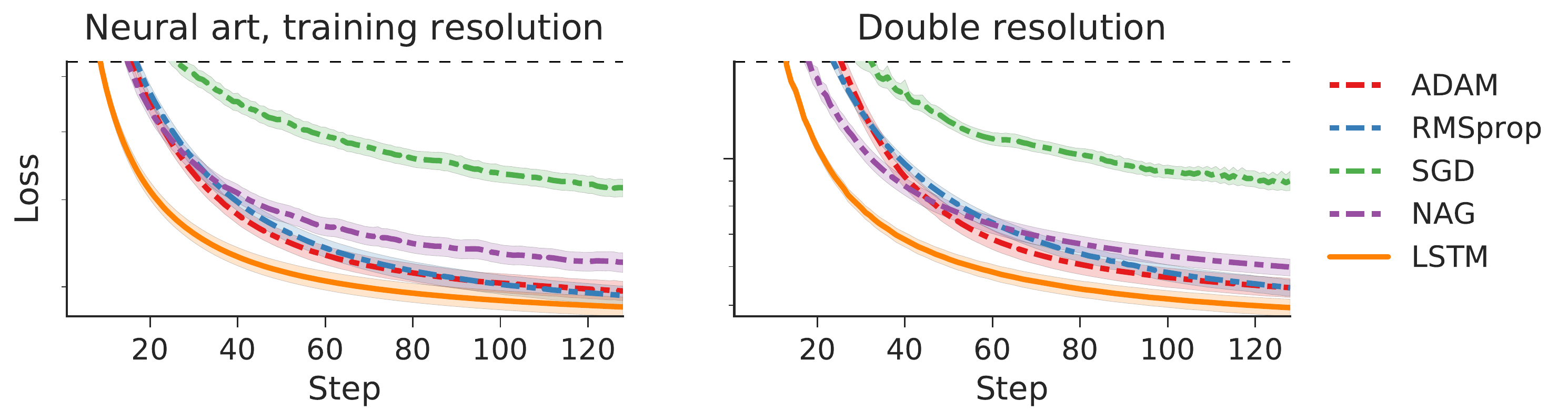}
\vspace{-0.1cm}
\caption{
Optimization curves for Neural Art. Content images come from the test set,
which was not used during the LSTM optimizer training. Note: the y-axis is in
log scale and we zoom in on the interesting portion of this plot.
\textbf{Left:} Applying the training style at the training resolution.
\textbf{Right:} Applying the test style at double the training resolution.}
\label{fig:neural-art}
\end{figure}

\begin{figure}
\centering
\includegraphics[width=0.46\linewidth]{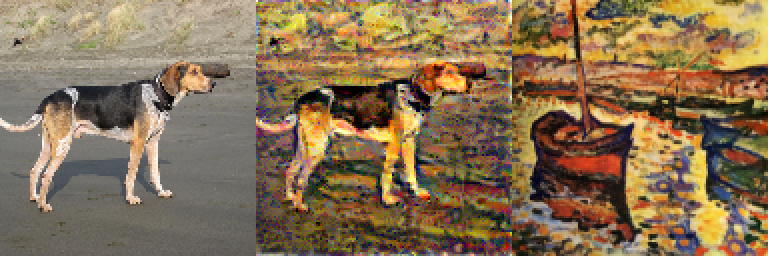}
\qquad
\includegraphics[width=0.46\linewidth]{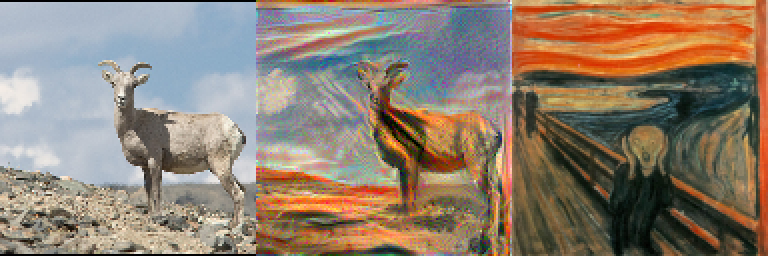}
\caption{
Examples of images styled using the LSTM optimizer.  Each triple consists of
the content image (left), style (right) and image generated by the LSTM
optimizer (center).  \textbf{Left:} The result of applying the training style
at the training resolution to a test image.  \textbf{Right:} The result of
applying a new style to a test image at double the resolution on which the
optimizer was trained.}
\label{fig:neural-art-examples}
\end{figure}

The recent work on artistic style transfer using convolutional networks, or
Neural Art \citep{gatys:2015}, gives a natural testbed for our method, since
each content and style image pair gives rise to a different optimization
problem.  Each Neural Art problem starts from a \emph{content image}, $c$, and
a \emph{style image}, $s$, and is given by
\begin{align*}
f(\theta) = \alpha \mathcal{L}_\mathrm{content}(c, \theta)
+ \beta \mathcal{L}_\mathrm{style}(s, \theta) + \gamma \calL_{\mathrm{reg}}(\theta)
\,
\end{align*}
The minimizer of $f$ is the \emph{styled image}.
The first two terms try to match the content and style of the styled image to that of their first argument, and the third term is a regularizer that encourages smoothness in the styled image.
Details can be found in \citep{gatys:2015}.


We train optimizers using only 1 style and 1800 content images taken from
ImageNet \citep{deng:2009}.  We randomly select 100 content images for testing
and 20 content images for validation of trained optimizers.  We train the
optimizer on 64x64 content images from ImageNet and one fixed style image.  We
then test how well it generalizes to a different style image and higher
resolution (128x128).  Each image was optimized for 128 steps and trained
optimizers were unrolled for 32 steps.  Figure~\ref{fig:neural-art-examples}
shows the result of styling two different images using the LSTM optimizer.  The
LSTM optimizer uses inputs preprocessing described in Appendix~\ref{sec:pre}
and no postprocessing. See Appendix~\ref{sec:neural-art-images} for additional images.

Figure~\ref{fig:neural-art} compares the performance of the LSTM optimizer to
standard optimization algorithms.  The LSTM optimizer outperforms all standard
optimizers if the resolution and style image are the same as the ones on which
it was trained.  Moreover, it continues to perform very well when both the
resolution and style are changed at test time.

Finally, in Appendix~\ref{sec:proposed-updates} we qualitatively examine the
behavior of the step directions generated by the learned optimizer.



\section{Conclusion}

We have shown how to cast the design of optimization algorithms as a learning
problem, which enables us to train optimizers that are specialized to
particular classes of functions.  Our experiments have confirmed that learned
neural optimizers compare favorably against state-of-the-art optimization
methods used in deep learning. We witnessed a remarkable degree of transfer,
with for example the LSTM optimizer trained on 12,288 parameter neural art
tasks being able to generalize to tasks with 49,152 parameters, different
styles, and different content images all at the same time. We observed similar
impressive results when transferring to different architectures in the MNIST
task.

The results on the CIFAR image labeling task show that the LSTM optimizers
outperform hand-engineered optimizers when transferring to datasets drawn from
the same data distribution.

{
\setlength{\bibsep}{0.4pt}
\renewcommand{\bibfont}{\small}
\titlespacing*{\section}{0pt}{0pt}{0pt}
\bibliographystyle{abbrvnat}
\bibliography{learning-to-learn}
}

\newpage
\appendix
\section{Gradient preprocessing}\label{sec:pre}

One potential challenge in training optimizers is that different input coordinates (i.e. the gradients w.r.t. different
optimizee parameters) can have very different magnitudes.
This is indeed the case e.g. when the optimizee is a neural network and different parameters
correspond to weights in different layers.
This can make training an optimizer difficult, because neural networks
naturally disregard small variations in input signals and concentrate on bigger input values.

To this aim we propose to preprocess the optimizer's inputs.
One solution would be to give the optimizer $\left(\log(|\nabla|),\,\operatorname{sgn}(\nabla)\right)$ as an input, where
$\nabla$ is the gradient in the current timestep.
This has a problem that $\log(|\nabla|)$ diverges for $\nabla \rightarrow 0$.
Therefore, we use the following preprocessing formula 
\begin{align*}
    \nabla^k &\rightarrow
\begin{cases}
    \left(\frac{\log(|\nabla|)}{p}\,, \operatorname{sgn}(\nabla)\right)& \text{if } |\nabla| \geq e^{-p}\\
    (-1, e^p \nabla)             & \text{otherwise}
\end{cases}
\end{align*}
where $p>0$ is a parameter controlling how small gradients are disregarded (we use $p=10$ in all our experiments).

We noticed that just rescaling all inputs by an appropriate constant instead also works fine, but
the proposed preprocessing seems to be more robust and gives slightly better results on some problems.





\section{Visualizations}
\label{sec:proposed-updates}

Visualizing optimizers is inherently difficult because their proposed updates
are functions of the full optimization trajectory.  In this section we try to
peek into the decisions made by the LSTM optimizer, trained on the neural art
task.

\paragraph{Histories of updates}

\begin{figure}
\centering
\includegraphics[width=\linewidth]{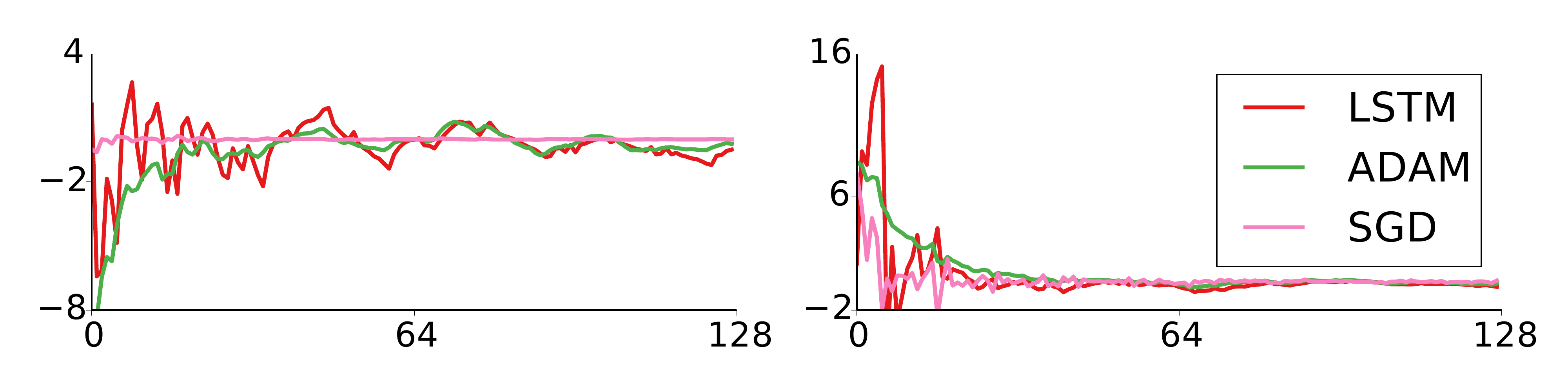}
\caption{Updates proposed by different optimizers at different optimization
steps for two different coordinates.}
\label{fig:analysis-a}
\end{figure}

We select a single optimizee parameter (one color channel of one pixel in the
styled image) and trace the updates proposed to this coordinate by the LSTM
optimizer over a single trajectory of optimization.  We also record the updates
that would have been proposed by both SGD and ADAM if they followed the same
trajectory of iterates.  Figure~\ref{fig:analysis-a} shows the trajectory of
updates for two different optimizee parameters.  From the plots it is clear
that the trained optimizer makes bigger updates than SGD and ADAM.  It is also
visible that it uses some kind of momentum, but its updates are more noisy than
those proposed by ADAM which may be interpreted as having a shorter time-scale
momentum.

\paragraph{Proposed update as a function of current gradient}

Another way to visualize the optimizer behavior is to look at the proposed
update $g_t$ for a single coordinate as a function of the current gradient
evaluation $\nabla_t$.  We follow the same procedure as in the previous
experiment, and visualize the proposed updates for a few selected time steps.

\begin{figure}
\centering
\includegraphics[width=\linewidth]{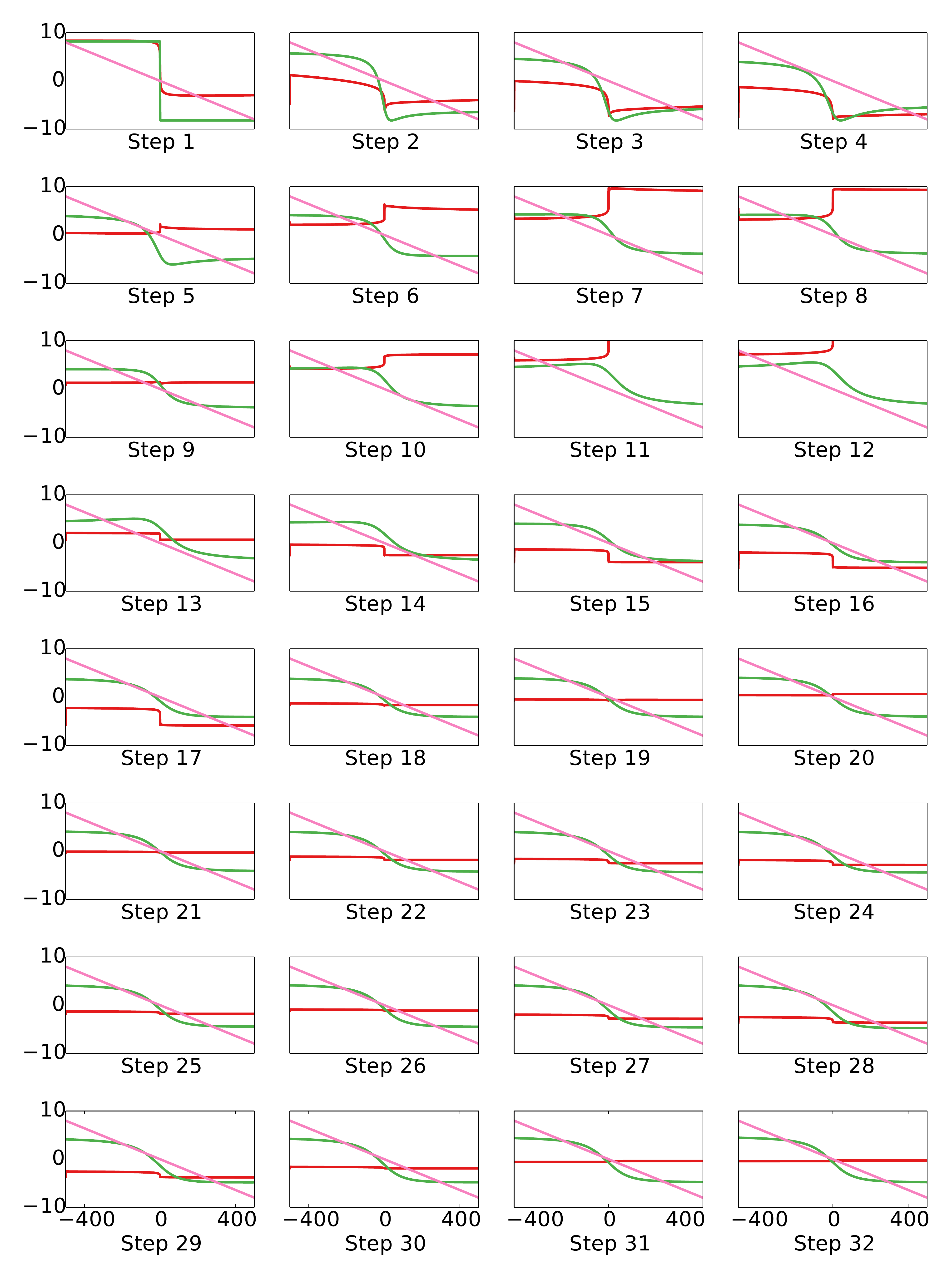}
\caption{The proposed update direction for a single coordinate over 32 steps.}
\label{fig:coordinate-1}
\end{figure}

\begin{figure}
\centering
\includegraphics[width=\linewidth]{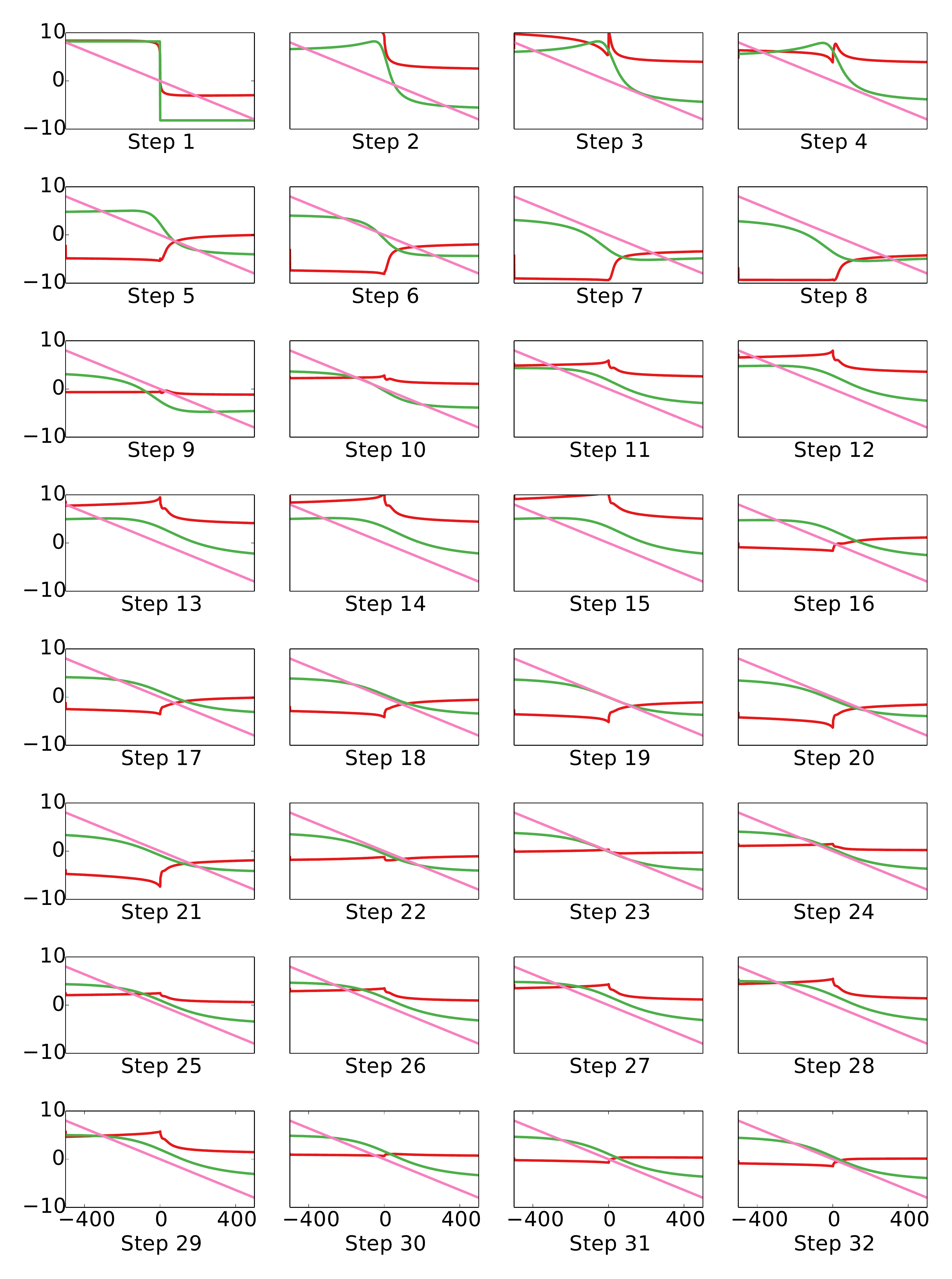}
\caption{The proposed update direction for a single coordinate over 32 steps.}
\label{fig:coordinate-2}
\end{figure}

\begin{figure}
\centering
\includegraphics[width=\linewidth]{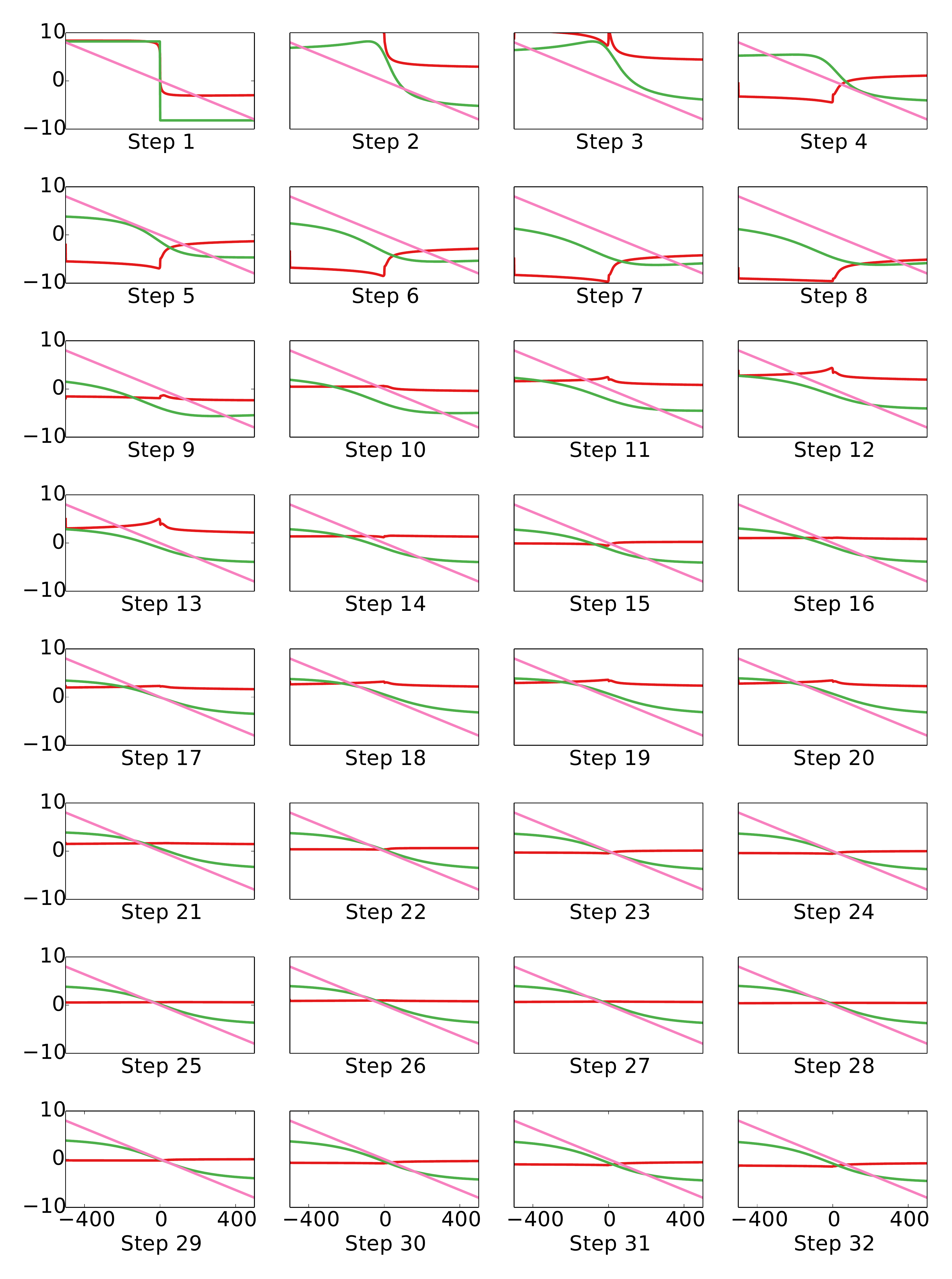}
\caption{The proposed update direction for a single coordinate over 32 steps.}
\label{fig:coordinate-3}
\end{figure}

These results are shown in
Figures~\ref{fig:coordinate-1}--\ref{fig:coordinate-3}.  In these plots, the
$x$-axis is the current value of the gradient for the chosen coordinate, and
the $y$-axis shows the update that each optimizer would propose should the
corresponding gradient value be observed.  The history of gradient observations
is the same for all methods and follows the trajectory of the LSTM optimizer.

The shape of this function for the LSTM optimizer is often step-like, which is
also the case for ADAM.  Surprisingly the step is sometimes in the opposite
direction as for ADAM, i.e.\ the bigger the gradient, the bigger the update.

\section{Neural Art}
\label{sec:neural-art-images}

Shown below are additional examples of images styled using the LSTM optimizer.
Each triple consists of the content image (left), style (right) and image
generated by the LSTM optimizer (center).

\vspace{1em}

\includegraphics[width=0.46\linewidth]{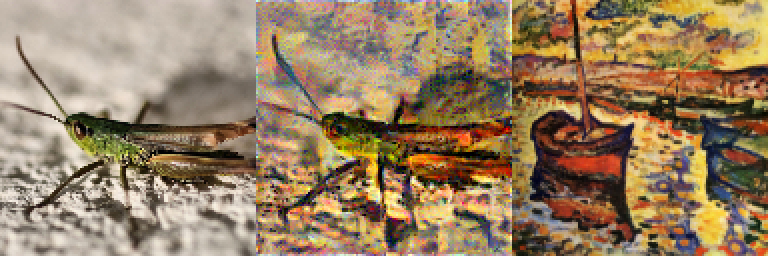}
\qquad
\includegraphics[width=0.46\linewidth]{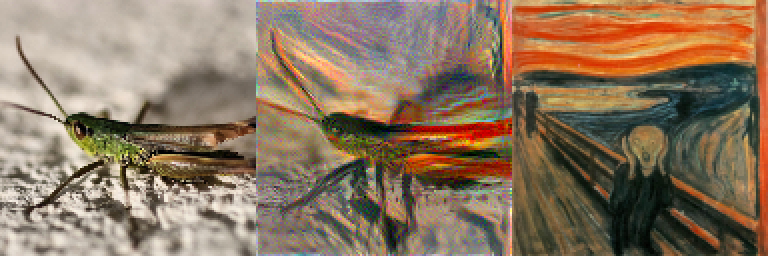}
\\
\includegraphics[width=0.46\linewidth]{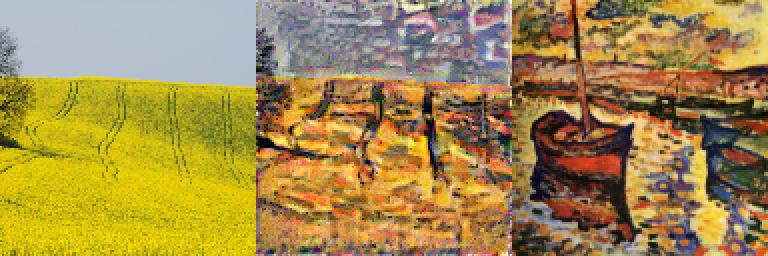}
\qquad
\includegraphics[width=0.46\linewidth]{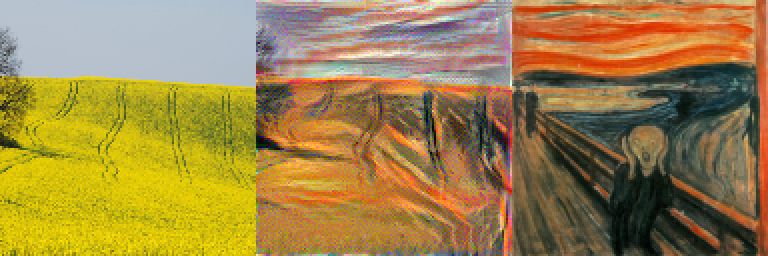}
\\
\includegraphics[width=0.46\linewidth]{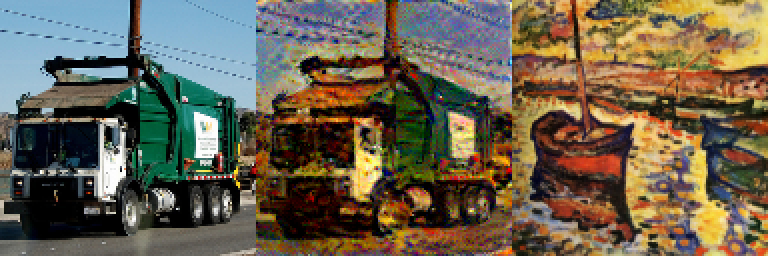}
\qquad
\includegraphics[width=0.46\linewidth]{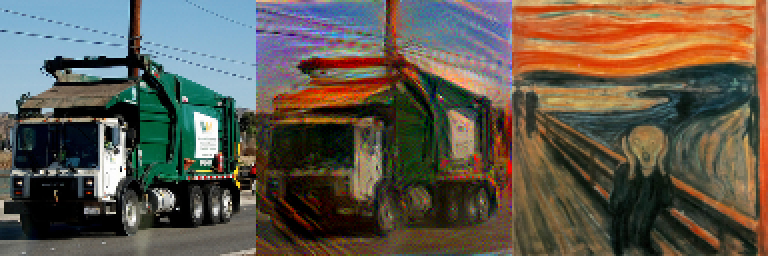}
\\
\includegraphics[width=0.46\linewidth]{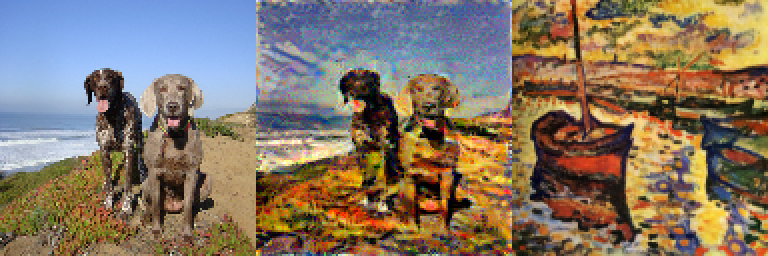}
\qquad
\includegraphics[width=0.46\linewidth]{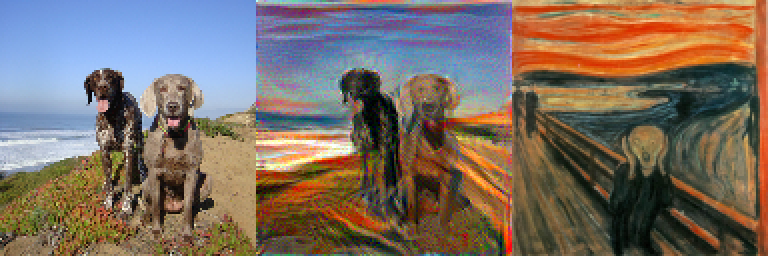}

\section{Information sharing between coordinates}
\label{sec:ntm}

In previous sections we considered a coordinatewise architecture, which
corresponds by analogy to a learned version of RMSprop or ADAM.  Although
diagonal methods are quite effective in practice, we can also consider learning
more sophisticated optimizers that take the correlations between coordinates
into effect. To this end, we introduce a mechanism allowing different LSTMs to
communicate with each other.

\subsection*{Global averaging cells}

The simplest solution is to designate a subset of the cells in each LSTM layer
for communication. These cells operate like normal LSTM cells, but their
outgoing activations are averaged at each step across all coordinates.  These
\emph{global averaging cells} (GACs) are sufficient to allow the networks to
implement L2 gradient clipping \citep{bengio:2013} assuming that each LSTM can
compute the square of the gradient. This architecture is denoted as an LSTM+GAC
optimizer.

\subsection*{NTM-BFGS optimizer}

We also consider augmenting the LSTM+GAC architecture with an external memory
that is shared between coordinates. Such a memory, if appropriately designed
could allow the optimizer to learn algorithms similar to (low-memory)
approximations to Newton's method, e.g.\ (L-)BFGS \citep[see][]{nocedal:2006}.
The reason for this interpretation is that such methods can be seen as a set of
independent processes working coordinatewise, but communicating through the
inverse Hessian approximation stored in the memory.  We designed a memory
architecture that, in theory, allows the network to simulate (L-)BFGS,
motivated by the approximate Newton method BFGS, named for Broyden, Fletcher,
Goldfarb, and Shanno.  We call this architecture an \emph{NTM-BFGS optimizer},
because its use of external memory is similar to the Neural Turing Machine
\citep{graves:2014}.  The pivotal differences between our construction and the
NTM are (1) our memory allows only low-rank updates; (2) the controller
(including read/write heads) operates coordinatewise.

In BFGS an explicit estimate of the full (inverse) Hessian is built up from the sequence of observed gradients.
We can write a skeletonized version of the BFGS algorithm, using $M_t$ to represent the inverse Hessian approximation at iteration $t$, as follows
\begin{align*}
g_t &= \operatorname{read}(M_t, \theta_t) \\
\theta_{t+1} &= \theta_{t} + g_t \\
M_{t+1} &= \operatorname{write}(M_t, \theta_t, g_t)
\,.
\end{align*}
Here we have packed up all of the details of the BFGS algorithm into the
suggestively named $\operatorname{read}$ and $\operatorname{write}$ operations,
which operate on the inverse Hessian approximation $M_t$.  In BFGS these
operations have specific forms, for example $\operatorname{read}(M_t, \theta_t)
= -M_t\nabla f(\theta_t)$ is a specific matrix-vector multiplication and
the BFGS write operation corresponds to a particular low-rank update of $M_t$.

In this work we preserve the structure of the BFGS updates, but discard their
particular form. More specifically the $\operatorname{read}$ operation remains
a matrix-vector multiplication but the form of the vector used is learned.
Similarly, the $\operatorname{write}$ operation remains a low-rank update, but the
vectors involved are also learned. Conveniently, this structure of interaction with
a large dynamically updated state corresponds in a fairly direct way to the
architecture of a Neural Turing Machine (NTM), where $M_t$ corresponds to the
NTM memory \citep{graves:2014}.

\begin{figure}
\centering
\includegraphics[width=0.54\linewidth]{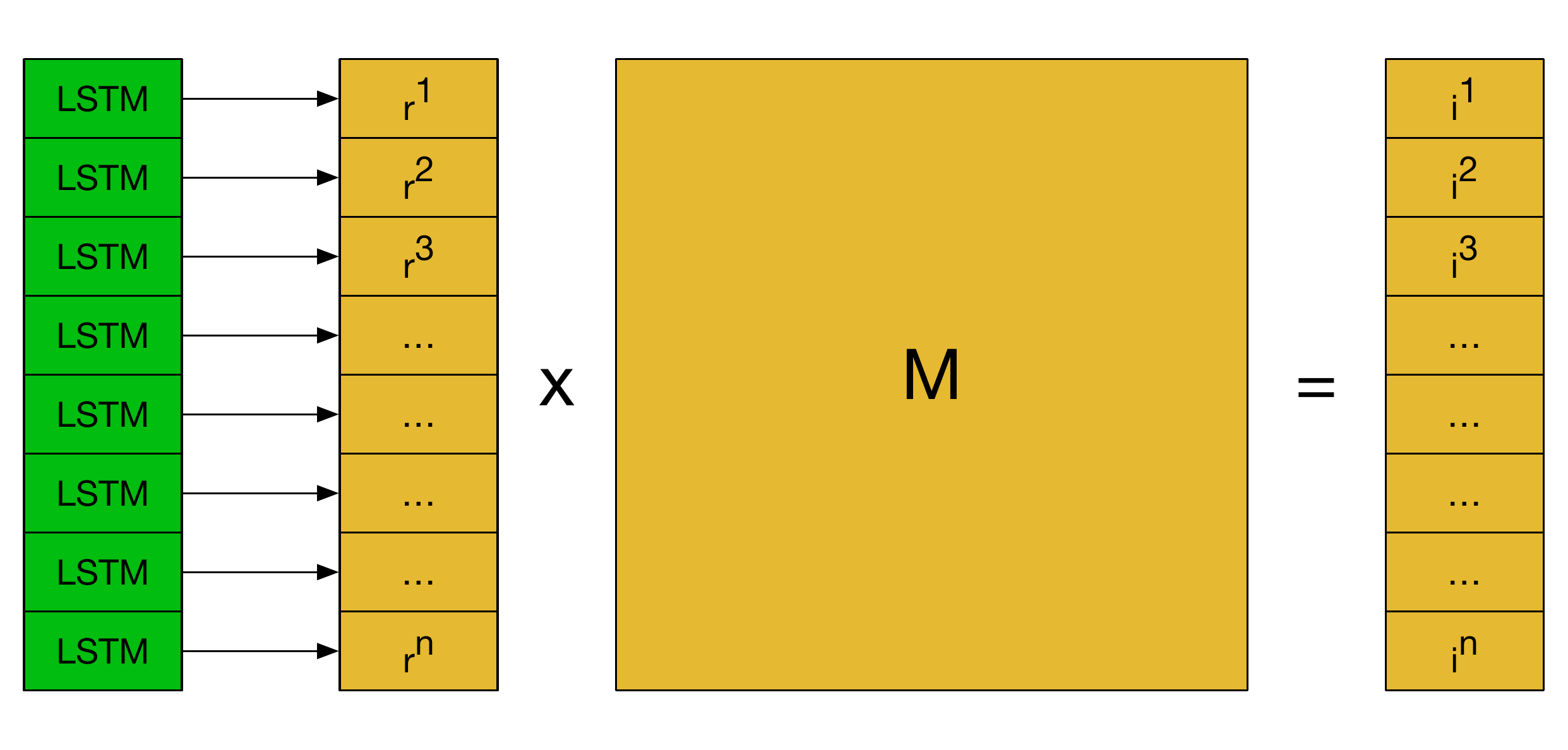}
\qquad
\includegraphics[width=0.38\linewidth]{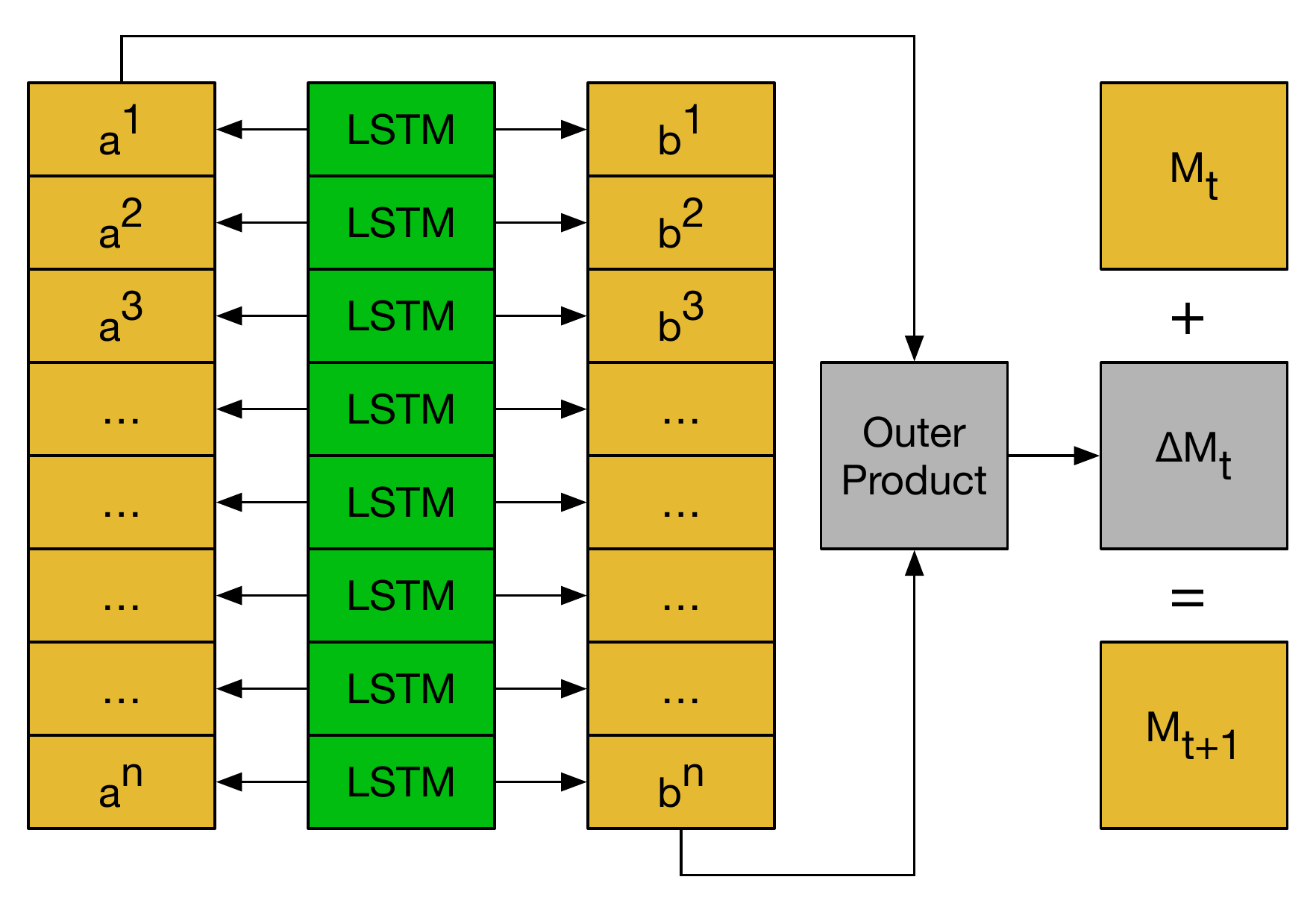}
\caption{\textbf{Left:} NTM-BFGS read operation. \textbf{Right:} NTM-BFGS write operation.}
\label{fig:ntm-bfgs}
\end{figure}

\begin{figure}
\centering
\includegraphics[width=0.6\linewidth,valign=t]{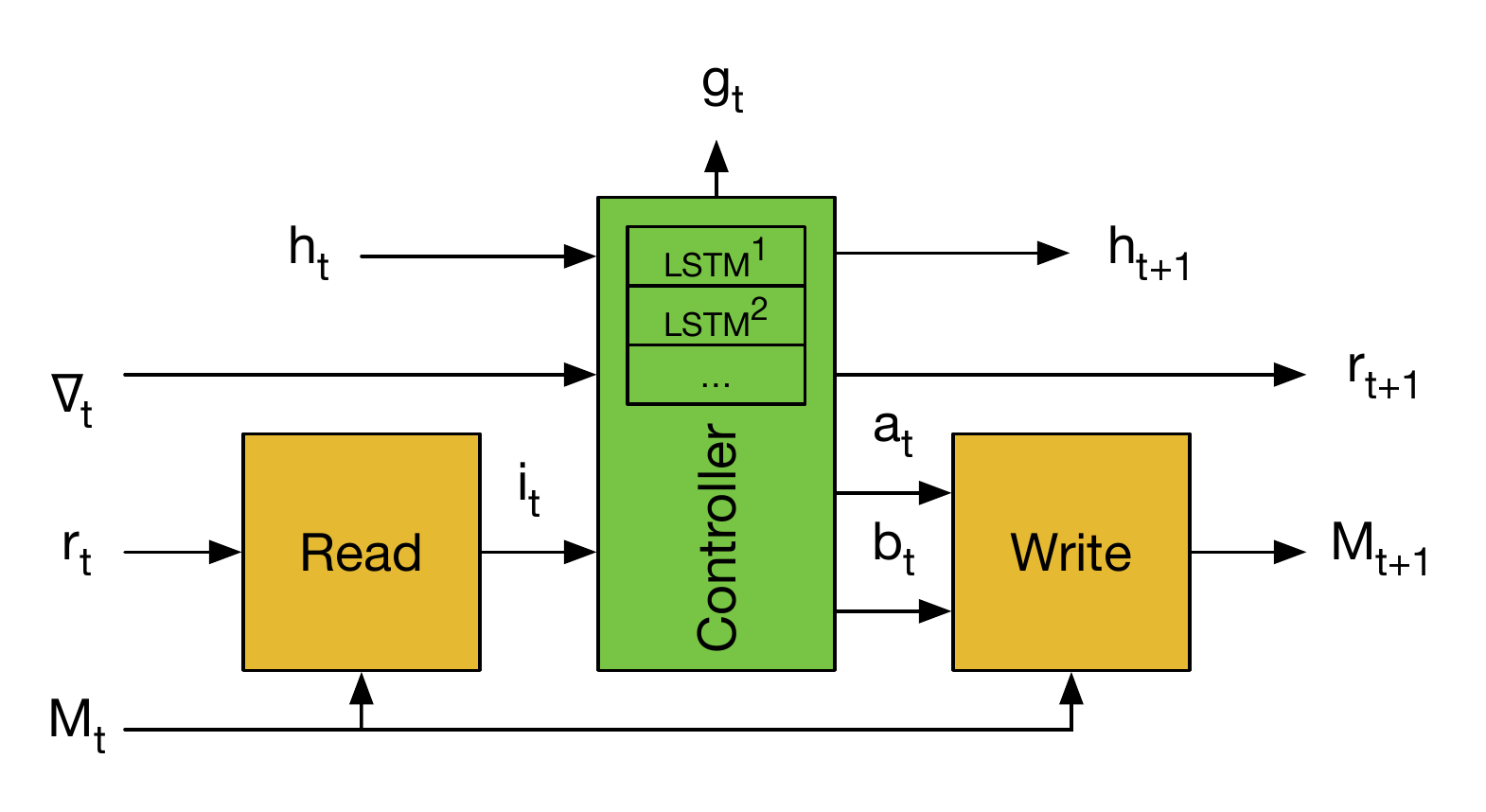}
\qquad
\includegraphics[width=0.32\linewidth,valign=t]{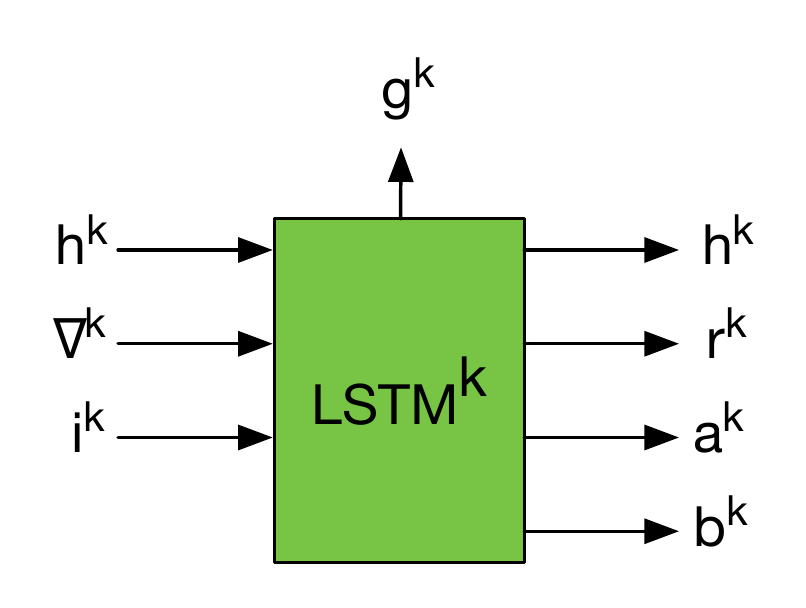}
\caption{\textbf{Left:} Interaction between the controller and the external
memory in NTM-BFGS. The controller is composed of replicated coordinatewise
LSTMs (possibly with GACs), but the read and write operations are global across all
coordinates.
\textbf{Right:}  A single LSTM for the $k$th coordinate in the NTM-BFGS
controller. Note that here we have dropped the time index $t$ to simplify
notation.}
\label{fig:ntm-controller}
\end{figure}

Our NTM-BFGS optimizer uses an LSTM+GAC as a controller;
however, instead of producing the update directly we attach one or
more read and write heads to the controller.
Each read head produces a read
vector $r_{t}$ which is combined with the memory to produce a read result $i_t$ which is fed back into the controller at the following time
step.  Each write head produces two outputs, a left write vector $a_t$
and a right write vector $b_t$. The two write vectors are used to update the
memory state by accumulating their outer product. The read and write operation for a single head is
diagrammed in Figure~\ref{fig:ntm-bfgs} and the way read and write heads are attached to the controller
is depicted in Figure~\ref{fig:ntm-controller}.

In can be shown that NTM-BFGS with one read head and 3 write heads can simulate inverse Hessian BFGS assuming that the controller
can compute arbitrary (coordinatewise) functions and have access to 2 GACs.

\subsection*{NTM-L-BFGS optimizer}

In cases where memory is constrained we can follow the example of
L-BFGS and maintain a low rank approximation of the
full memory (\emph{vis}.\ inverse Hessian).
The simplest way to do this is to store a sliding history of the left and right write vectors,
allowing us to form the matrix vector multiplication required by the read operation efficiently.




\end{document}